\pdfoutput=1
\documentclass[11pt]{article}
\usepackage{EMNLP2023}
\usepackage{times}
\usepackage{latexsym}
\usepackage[T1]{fontenc}
\usepackage[utf8]{inputenc}
\usepackage{microtype}
\usepackage{inconsolata}
\usepackage[most]{tcolorbox} 
\usepackage{graphicx}
\usepackage{amsmath, amssymb}
\usepackage{booktabs}
\usepackage{tabularx, array, adjustbox}
\usepackage{float}
\usepackage{caption}
\usepackage{hyperref}
\usepackage{fontawesome}
\usepackage{pifont}
\usepackage{stfloats}

\newcolumntype{L}[1]{>{\raggedright\arraybackslash}p{#1}}
\newcolumntype{C}[1]{>{\centering\arraybackslash}p{#1}}

\title{Deep Associations, High Creativity: A Simple yet Effective Metric for Evaluating Large Language Models}

\author{Ziliang Qiu \\
University of Illinois, Urbana-Champaign \\
Beijing Normal University \\
\texttt{ziliang6@illinois.edu} \\
\And
Renfen Hu~\ding{41} \\
Beijing Normal University \\
\texttt{irishu@mail.bnu.edu.cn} \\}

\begin{document}
\maketitle
\begin{abstract}
The evaluation of LLMs' creativity represents a crucial research domain, though challenges such as data contamination and costly human assessments often impede progress. Drawing inspiration from human creativity assessment, we propose PACE, asking LLMs to generate \textbf{P}arallel \textbf{A}ssociation \textbf{C}hains to \textbf{E}valuate their creativity. PACE minimizes the risk of data contamination and offers a straightforward, highly efficient evaluation, as evidenced by its strong correlation with Chatbot Arena Creative Writing rankings (Spearman's $\rho = 0.739$, $p < 0.001$) across various proprietary and open-source models. A comparative analysis of associative creativity between LLMs and humans reveals that while high-performing LLMs achieve scores comparable to average human performance, professional humans consistently outperform LLMs. Furthermore, linguistic analysis reveals that both humans and LLMs exhibit a trend of decreasing concreteness in their associations, and humans demonstrating a greater diversity of associative patterns.\footnote{Our code and data are publicly available at \url{https://github.com/ziliang6/PACE}}
\end{abstract}

\section{Introduction}
Developing creative artificial intelligence and boosting co-creativity remain central goals in AI research \cite {rafner2023creativity, franceschelli2024creativity, lee2024empirical}. 
% Current research conducts diverse creativity-based tasks to evaluate the creative capabilities of Large Language Models (LLMs), aiming to understand their potential and limitations
Recent studies evaluate the creative capabilities of large language models (LLMs) through diverse tasks, aiming to understand their strengths and limitations \cite{tian2023macgyver, atmakuru2024cs4, si2024can}.

\begin{figure}[t]
\centering
\vspace{-20pt}
\includegraphics[width=1\columnwidth]{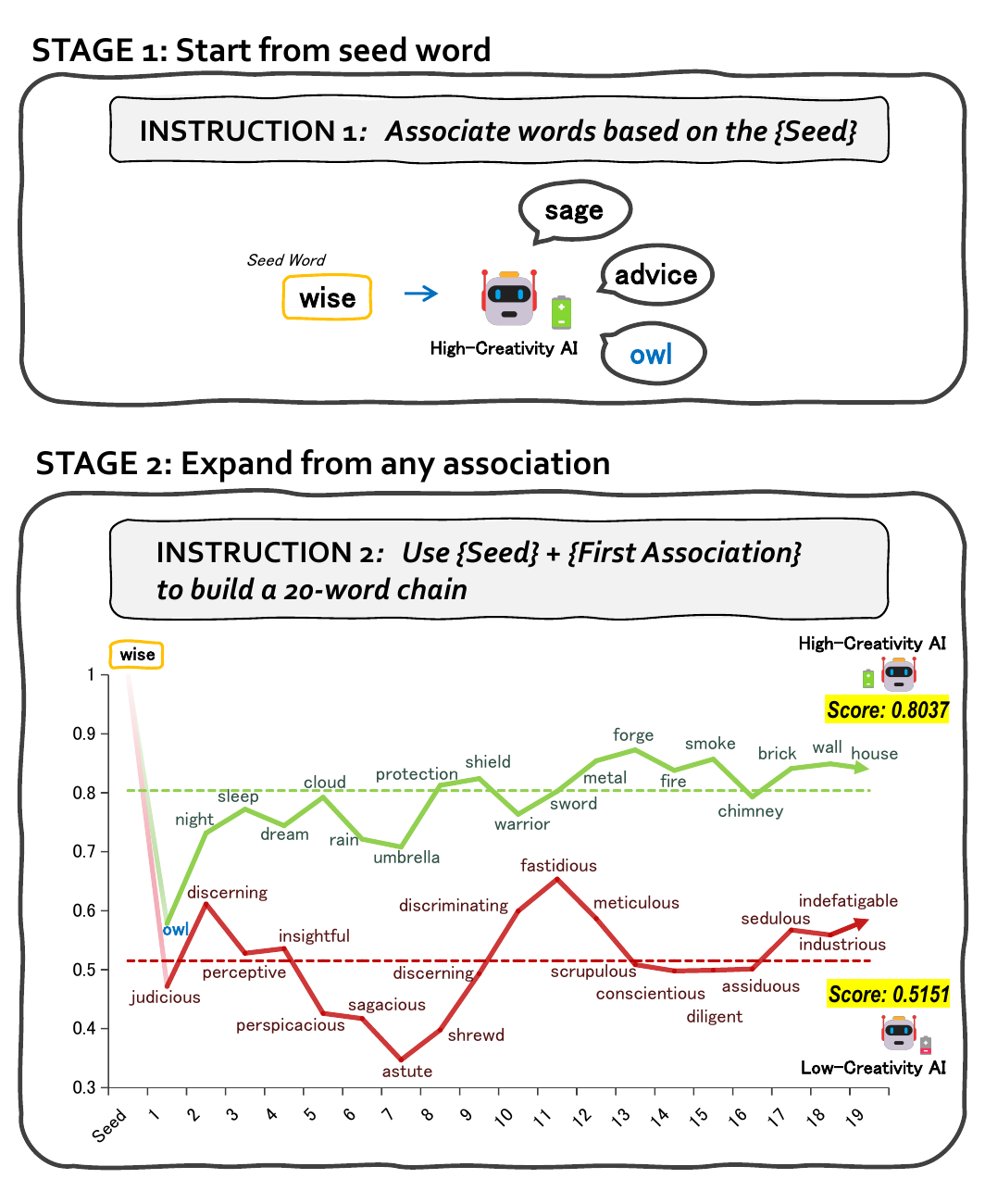}
\vspace{-20pt}
\caption{PACE evaluation process: For each seed word, three 20-word association chains are generated and their association distances are averaged to obtain the seed score. The model's creativity score is calculated by averaging all seed scores.}
\label{fig:structure}
\end{figure}

However, data contamination, a prominent issue in current LLM evaluations, may compromise the reliability of conclusions \cite{sainz2023nlp, xu2024benchmark, lu2024ai}. Moreover, unlike tasks with definitive answers, establishing frameworks to evaluate creativity poses unique challenges, particularly due to its complex nature \cite{rafner2023creativity, ivcevic2024artificial} and the subjective and time-consuming process of human scoring \cite{olson2021naming, organisciak2023beyond, lu2024benchmarking}.

In light of these issues, this study draws inspiration from established psycholinguistic measures of human creativity and introduces PACE (Parallel Association Chain Evaluation), a highly efficient framework to evaluate LLMs. As shown in Figure~\ref{fig:structure}, this approach requires no human-annotated data and enables automatic and reliable scoring. Associative evaluation lies at the core of human creativity research \cite{mednick1968remote, olson2021naming, beaty2023associative}. The theory of associative creativity posits that individuals with higher creative capacity are more likely to generate unconventional connections, enabling them to link disparate concepts and produce original ideas \cite{mednick1962associative, merseal2023free}. As for LLMs, measuring associative distance efficiently assesses their capacity for creative association, reflecting their ability to move beyond surface co-occurrence patterns and tap into deeper, less common semantic links that underlie genuine creativity \cite{yao2022wordties, abramski2024llm}.

Our results demonstrate a strong correlation between PACE and Arena Creative Writing rankings ($\rho = 0.739$, $p < 0.001$), as well as other LLM leaderboards, through testing a wide range of open-source and closed-source models of varying capabilities. We further compare associative creativity between humans and LLMs, showing that state-of-the-art models perform comparably to general human groups but still fall short of professionals. Linguistic analysis reveals that both produce associations with decreasing concreteness; however, human associations are generally more abstract and exhibit greater diversity in association types.

\begin{figure*}[t]
\centering
\vspace{-10pt}
\includegraphics[width=0.8\textwidth]{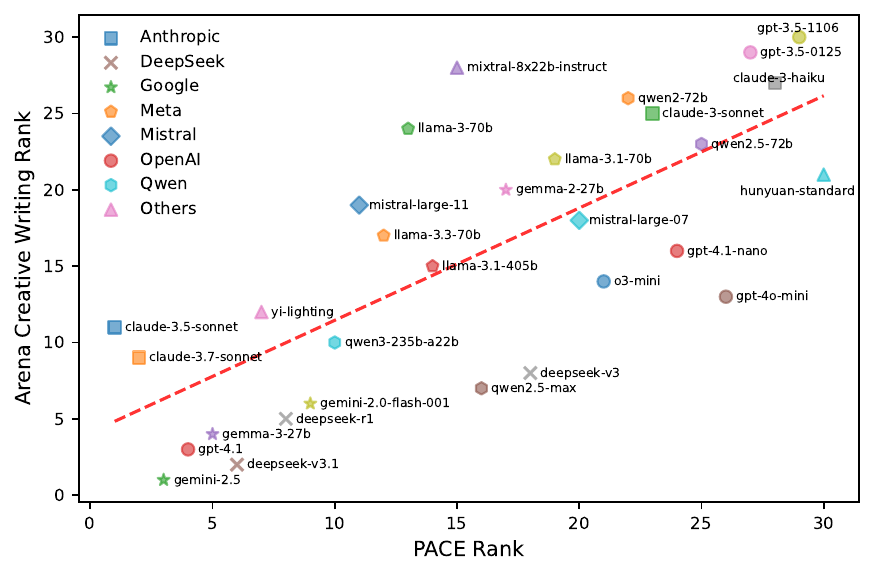}
\vspace{-10pt}
\caption{Comparison of model rankings according to PACE and Arena Creative Writing. Each point represents a language model, where different release versions of the same model are treated as separate variants, with the x-axis showing the PACE rank (based on association distance) and the y-axis showing the Arena Creative Writing rank. The red dashed line indicates the Spearman rank correlation fit ($\rho = 0.739$, $p < 0.001$). Claude-3.5-Sonnet achieves the highest PACE ranking among the evaluated models.}
\label{fig:mean_rank}
\vspace{-10pt}
\end{figure*}

\section{Related Work}
\subsection{Evaluating LLMs' Creativity}
LLMs have demonstrated remarkable capabilities in diverse creative tasks, leading researchers to design increasingly complex evaluations that explore their potential for creative writing \cite{doshi2024generative, tian2024large, walsh2024does}, scientific hypothesis generation \cite{si2024can, tong2024automating}, and co-creativity with human \cite{dell2023navigating, ashkinaze2024ai, boussioux2024crowdless}. 
% Hu: code generation \cite{delorenzo2024creativeval, lu2024benchmarking} 不太适合作为创造力任务，故删去

% In creativity evaluations involving open-ended questions where single correct answers are neither expected nor desired, human evaluation provides reliable preference data but presents significant challenges in implementation and reproducibility. This complexity is exemplified in a recent study by \citet{tian2023macgyver}, which developed constrained real-world questions designed to stimulate unconventional thinking in both human subjects and AI models. Their findings identified a potential limitation: human evaluators frequently disagreed in their assessments, primarily due to varying levels of question comprehension. To address these challenges in creativity measurement, researchers have increasingly adopted LLM-as-judge approaches \citep{organisciak2023beyond,raz2024open}. Nevertheless, this alternative methodology introduces distinct complications, particularly the potential emergence of systematic "LLM preferences," making LLM-based evaluation an inherently challenging research domain.
In creativity evaluations with open-ended questions, human assessment offers reliable preference data but poses significant challenges for implementation and reproducibility. \citet{tian2023macgyver} introduced constrained real-world questions to stimulate unconventional thinking and found that human evaluators often disagreed in their judgments, largely due to varying levels of question comprehension.
To overcome these challenges, researchers have increasingly adopted LLM-as-judge approaches \citep{organisciak2023beyond,raz2024open}. However, concerns remain regarding their reliability and fairness~\citep{stureborg2024large, thakur2024judging}.

\subsection{Adapting Human Creativity Assessments for LLM Evaluation}
Creativity is central to human intelligence, making its assessment a fundamental topic in psychological research. Recently, numerous psychological assessments have been applied to evaluate the creativity of LLMs, including the Alternative Uses Task \cite{koivisto2023best, hubert2024current}, the Remote Associates Test \cite{alavi2023large}, and the Torrance Tests of Creative Thinking \cite{guzik2023originality, hubert2024current}. 

% The prevalent methodology across these studies involves using LLMs to complete psychological assessments and comparing their performance with human participants.
Most existing studies employ LLMs to complete psychological assessments and compare their performance to that of human participants.
However, since these creativity tests are widely available, they risk training data contamination. For example, GPT-3 could produce responses that directly replicated content from psychological journals and test manuals \cite{stevenson2022putting}. 
% Furthermore, psychological assessments designed for human cognition lack empirical validation when applied to machine creativity, weakening the correlation between model scores and practical performance—an issue current research has yet to address.
Moreover, psychological assessments designed for human cognition lack empirical validation when applied to machine creativity, thereby undermining the correlation between model scores and real-world performance — a concern that current research has yet to adequately address.

\section{Method}
\subsection{Parallel Word Association Chains}
\label{sec:parallel_word_association}
The ability to generate distant associations is a key indicator of creativity, as it reveals unconventional connections between concepts and ideas \cite{mednick1962associative, kenett2014investigating, zhang2023retrieval}. Similarly, advanced models are expected to capture multi-level semantics and identify deeper connections, enabling them to foster novel insights.

To systematically evaluate this capability, we present a two-phase approach inspired by human participant studies from \citet{gray2019forward}. The approach consists of: (1) eliciting three distinct associations from LLMs as secondary seed words, and (2) generating 20-word association chains that contain both primary and secondary seeds.\footnote{We choose a length of 20 words to align with the human participant data collected by \citet{gray2019forward}, allowing for fair comparisons between models and humans in Section~\ref{sec:comparison_between_human_and_llms}. Furthermore, we compare different chain lengths. As shown in Table~\ref{tab:spearman_correlations_length}, PACE scores at this length (20 words) exhibit the strongest correlations with existing leaderboards.}

Each association chain is generated independently to minimize mutual influence among the chains. Compared to single-chain association, this parallel approach improves the diversity of associative pathways, allowing a broader sampling of the model’s creative potential. For each independent chain, we apply a chain-of-thought prompting strategy to guide the model’s word associations\footnote{While multi-turn dialogue could also be used to elicit associations, both approaches have limitations: generating without conversational history often leads to redundant outputs, while providing the full conversational history introduces confounds such as long-context memory and coherence constraints inherent to multi-turn setups. Therefore, we adopt independent prompts to obtain interpretable and controlled measurements of creative associative capacity. Although this does not exactly mirror human reasoning processes, it helps avoid the issues associated with multi-turn association setups and also facilitates more straightforward computational evaluation.}, ensuring a structured yet flexible generation process. Prompts can be found in Appendix~\ref{sec:prompts}.

\subsection{Seed Words}
110 seed words are selected from the Intercontinental Dictionary Series (IDS, \citealp{ids}), a multilingual project representing universal concepts across languages. The IDS consists of 22 chapters, each corresponding to a distinct semantic domain, such as time, quantity, and motion. From each chapter, five seed words are chosen based on their frequencies in the COCA corpus~\citep{davies2008coca}, using five equally spaced frequency intervals to ensure balanced representation. This selection process combines semantic diversity and frequency variation to enable a comprehensive evaluation. 
For each model, three chains per seed yield 6,270 associated words. The complete list of seed words is provided in Appendix~\ref{sec:selected_seed_words}.

\subsection{Association Distance Metric} 
We measure the creativity score using the mean association distance. 
Each seed's score is derived by averaging the association distances of three chains, and the model's overall associative creativity is determined by averaging the scores of 110 seeds. See details in Appendix~\ref{sec:formula_for_association_distances}. We use FastText (crawl-300d-2m; \citealp{mikolov2018advances}) for computing cosine distance. Table~\ref{tab:spearman_results_with_different_embeddings} also reports results using alternative word embedding models.

\section{Experiments and Results}
\subsection{Models and Parameters} 
% Thirty models are selected from the Chatbot Arena Leaderboard, balancing different performance ranks and license types (commercial and open-source). The models range from rank 1 (Gemini-2.5-Pro) to rank 184 (GPT-3.5-turbo-1106) out of 234 total models (May 2025). To enable robust correlation analysis with other benchmarks that typically evaluate fewer models, at least 18 models are included \cite{bonett2000sample}. Multiple Qwen model versions and sizes are added to analyze the scale-performance relationship. The complete model list appears in Table~\ref{tab:selected_models}. Model responses are obtained via APIs using temperature 0, except for o3-mini (temperature fixed at 1), with default settings for other parameters.
Thirty models are selected from the Chatbot Arena Leaderboard, representing a balanced coverage of different performance ranks and license types (commercial and open-source). The selection spans from rank 1 (Gemini-2.5-Pro) to rank 184 (GPT-3.5-turbo-1106) out of 234 models as of May 2025. To enable robust correlation analysis with existing benchmark — which typically evaluate fewer models — at least 18 models were included in each evaluation \cite{bonett2000sample}. Additionally, we compared Qwen models of varying versions and sizes on PACE. The complete list of models is provided in Table~\ref{tab:selected_models}. Model responses were obtained via APIs with a temperature setting of 0, except for o3-mini (temperature fixed at 1). All other parameters were default.

\subsection{Correlation with Existing Benchmarks}
We select several representative benchmarks to validate our results, including the Chatbot Arena leaderboard (Arena Overall ranking and Arena Creative Writing ranking, hereafter Arena All and Arena CW, which rank models based on human voting preferences for anonymous models, \citealp{chiang2024chatbot}), MMLU-Pro (a more complex and challenging version of Massive Multitask Language Understanding, \citealp{wang2024mmlu}), LiveBench (releasing new questions regularly, \citealp{white2024livebench}), EQ-Bench (specifically its creative writing leaderboard, scored by LLMs, \citealp{paech2023eq}). For each leaderboard, we calculate the Spearman correlation between the models' ranks in PACE and their ranks in the respective leaderboard.

\begin{table}[!t]
\centering
\caption{Spearman rank correlation between model rankings of PACE and different benchmarks.}
\label{tab:mean_rank}
\resizebox{\columnwidth}{!}{%
\begin{tabular}{cccc}
\hline
\textbf{Leaderboard} & \textbf{Corr.} & \textbf{P-value} & \textbf{Models} \\
\hline
Arena All & 0.660*** & $< 0.001$ & 30 \\
Arena CW & \textbf{0.739***} & $< 0.001$ & 30 \\
MMLU-Pro & 0.505* & $< 0.05$ & 23 \\
LiveBench & 0.691** & $< 0.01$ & 19 \\        
EQ-Bench & 0.637** & $< 0.01$ & 18 \\
\hline
\multicolumn{4}{l}{\small{* $p < 0.05$, ** $p < 0.01$, *** $p < 0.001$}} \\
\end{tabular}%
}
\end{table}

\subsection{Results}
% \textbf{PACE Demonstrates Strong Correlations With LLM Creative Performance Rankings.} 
As illustrated in Table~\ref{tab:mean_rank}, \textbf{the Spearman rank correlations between PACE and various leaderboards are consistently moderate to strong}. Notably, PACE exhibits its highest correlation with Arena CW (0.739***), which is substantially higher than with Arena All (0.660***), indicating that PACE better captures creative capabilities than general performance. 
% the Spearman rank correlation between PACE and various leaderboards ranges from moderate to strong. Notably, correlation analysis also validates PACE's creativity focus: PACE correlates most strongly with Arena CW (0.739***), notably higher than Arena All (0.660***), demonstrating that PACE effectively captures creative capabilities rather than general performance. The robustness of these correlations is confirmed through bootstrap analysis and alternative embedding approaches, with detailed results detailed in Appendix~\ref{sec:validation_results}.

% This key differential confirms creativity as a distinct performance dimension.

% Beyond aggregate correlations, PACE effectively differentiates models from the same organization with similar structures. As shown in Figure~\ref{fig:mean_rank}, DeepSeek-V3.1 scored 0.763 (rank 6), DeepSeek-R1 scored 0.759 (rank 8), and DeepSeek V3 scored 0.748 (rank 19), demonstrating PACE's discriminative power. Additionally, we analyze various versions and sizes of the Qwen model family, which provides diverse open-source variants for comparison (see Appendix~\ref{sec:pace_qwen_models}).
In addition to its strong correlations with existing benchmarks, \textbf{PACE effectively differentiates between model variants} within the same series. As shown in Figure~\ref{fig:mean_rank}, DeepSeek-V3.1 scored 0.763 (rank 6), DeepSeek-R1 scored 0.759 (rank 8), and DeepSeek-V3 scored 0.748 (rank 19). We also compare different Qwen models to investigate the effects of model version and size on association distances. Newer model generations consistently achieve higher scores (e.g., Qwen-3 > Qwen-2.5 > Qwen-2), while within the same generation, larger models tend to perform better. These results demonstrate that PACE is sensitive to subtle differences among models (see Appendix~\ref{sec:pace_qwen_models}).

\begin{figure}[t!]
    \centering
    \includegraphics[scale=0.5]{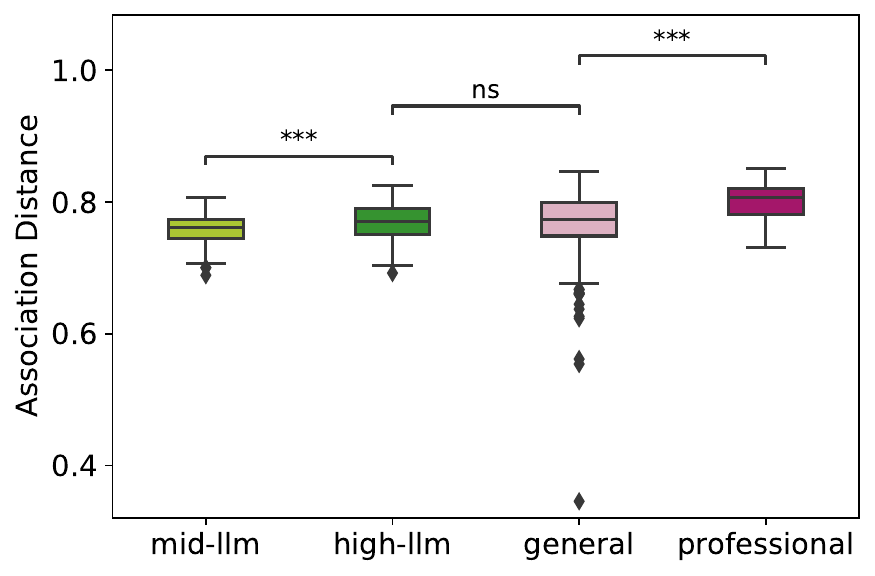}
    \vspace{-10pt} 
    \caption{Comparison of association distances between humans and LLMs. Using human data from \citet{gray2019forward}, results show that \textbf{high-performing LLMs match average human performance}, but fall short of professional humans. }
    \label{fig:association-distance-comparison}
\end{figure}

\section{Comparison between Humans and Models}
\label{sec:comparison_between_human_and_llms}

\subsection{Associative Creativity}
We compare human and LLM performance on associative creativity tests across four distinct groups. Human data from \citet{gray2019forward} includes a demographically representative sample of American adults (the \textit{general} group) and professional actors with higher creative abilities (the \textit{professional} group). For LLMs, we evaluate \textit{high-llm} models (top 20 on Arena leaderboard) and \textit{mid-llm} models (ranked around position 75 of 234 total). All models are tested using identical seed word prompts as those applied in the human studies. Details of experimental settings are presented in Appendix~\ref{sec:app_comparison_between_human_and_llms}.

\textbf{Current leading LLMs match average human creativity.} As shown in Figure~\ref{fig:association-distance-comparison}, high-performing models demonstrate comparable performance to \textit{general} human groups, with no significant difference observed (Welch's t-test: \(t = 0.644, \, p = 0.52 \)). This contrast with previous studies that reported significantly lower model performance compared to human participants \cite{wenger2025we}. Furthermore, high-performing models demonstrate significantly superior performance compared to mid-performing models (\( t = 3.781, \, p < 0.001 \)).

\begin{figure}[!t]
    \centering
    \includegraphics[scale=0.5]{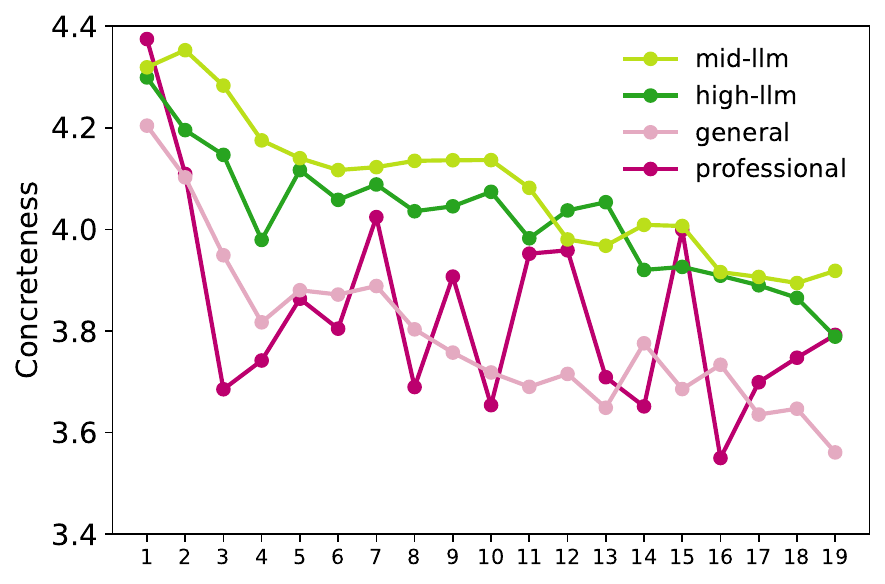}
    \vspace{-10pt} 
    \caption{Average concreteness scores across chain positions for different groups. All groups show declining concreteness, with models exhibiting higher concreteness than humans.}
    \label{fig:average_concrtness_in_position}
\end{figure}

\textbf{Best-performing human still outperforms LLMs.} Both the overall group scores (\( t = 6.152, \, p < 0.001 \)) and the maximum values (Human\textsubscript{max}=0.8501, Model\textsubscript{max}=0.8251) indicate that the top-performing humans still surpass the best LLMs, consistent with previous findings~\cite{koivisto2023best}. 
% show that the best-performing humans still outperform the best LLMs in agreement with previous research \cite{koivisto2023best}.
Moreover, a significant difference is observed between the professional group and all other groups, underscoring the unique value of human creativity~\cite{rafner2023creativity, lee2024empirical, boussioux2024crowdless}. 
In contrast, LLMs exhibit greater consistency in minimum performance (Human\textsubscript{min} = 0.3457; Model\textsubscript{min} = 0.6888), highlighting their potential as reliable co-creativity tools for generating consistent solutions \cite{dell2023navigating, jia2024and, lee2024empirical, ashkinaze2024ai}.

\subsection{Associative Patterns}
% We further compare the patterns of association between humans and LLMs from two aspects: trend of associations and type between associations. 
We further compare the patterns of association between humans and LLMs from two perspectives: the overall trends in associations and the types of associations observed.

\textbf{Trends of associations.} As shown in Figure~\ref{fig:average_concrtness_in_position}, both humans and LLMs exhibit a decreasing trend in concreteness as the chain develops. However, the models consistently demonstrate higher average concreteness scores compared to humans at each step. This suggests that LLMs tend to rely more on concrete concepts rather than abstract ones, whereas humans are more inclined toward abstract cognition as they progress through the association chain. Furthermore, while both LLMs and the general human population show a relatively steady decline in concreteness, professionals exhibit greater variability, suggesting more frequent transitions between concrete and abstract associations \cite{kenett2014investigating, zhang2023retrieval}. Fixed effects regression analysis confirmed significant declining trends in concreteness across all groups (all $p < 0.01$, see Appendix~\ref{sec:prediction_of_concreteness_using_embedding_models}), with LLMs showing higher baseline concreteness than humans.

\begin{figure}[ht]
    \centering
    \includegraphics[scale=0.5]{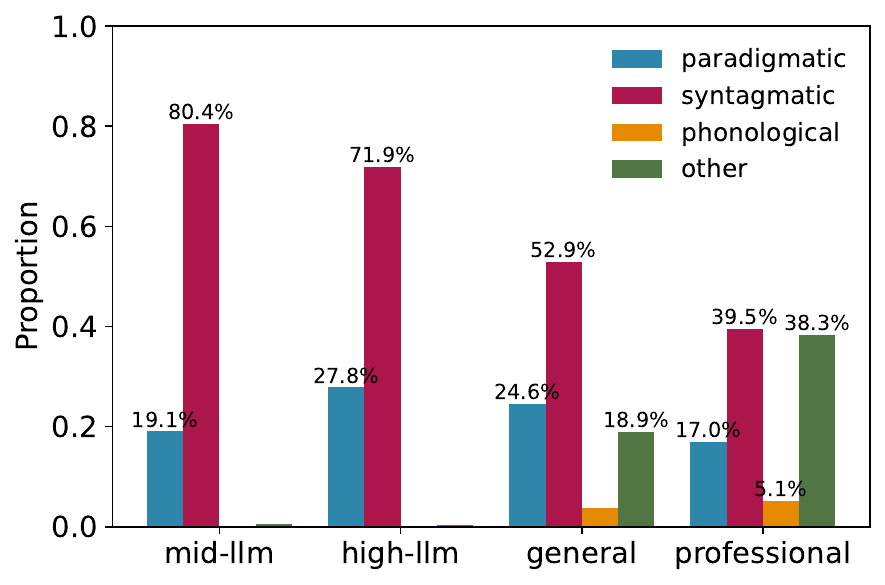}
    \vspace{-10pt} 
    \caption{Types of associations within chains, categorized according to the association type framework described by \citet{nissen2006word}. Details are provided in Appendix~\ref{sec:prediction_of_concreteness_using_embedding_models}.}
    \label{fig:association_type}
\end{figure}

\textbf{Associations types.} 
% Similar to humans, LLMs demonstrate a stronger tendency to generate syntagmatic associations (words that co-occur in sequences, like "dog" → "bark") compared to paradigmatic associations (words that can substitute for each other, like "dog" → "cat"). However, human associate more diversely, generating non-semantic relationships such as phonological connections. Moreover, association patterns among professionals show a tendency in "other" type of association, suggesting that creative individuals tend to form associations based on personal experiences rather than common linguistic patterns.
Like humans, LLMs show a stronger tendency to produce syntagmatic associations (e.g., \textit{dog} → \textit{bark}) than paradigmatic associations (e.g., \textit{dog} → \textit{cat}). However, humans demonstrate greater diversity in their associations, often generating non-semantic links such as phonological connections. Notably, professionals are more likely to produce \textit{other} types of associations, suggesting that creative individuals often draw on personal experiences rather than relying solely on common linguistic patterns.

\section{Conclusions}
% We propose PACE as a methodological framework to benchmark LLMs' creative associative abilities. Our findings demonstrate a strong and significant Spearman's rank correlation between PACE and several established leaderboards, e.g. $\rho = 0.739$ with Arena CW. Our results prove that measuring associative distance provides an efficient way to assess LLMs' capacity for creative association, reflecting their ability to move beyond surface co-occurrence patterns and tap into deeper, less common semantic links that underlie genuine creativity.

We propose PACE, a benchmark for evaluating the creative potential of LLMs based on parallel association chains. Compared to existing methods, PACE avoids training data contamination and offers a simple, scalable framework that greatly reduces manual evaluation costs. Experimental results demonstrate a strong and significant Spearman’s rank correlation between PACE and several established leaderboards (e.g., $\rho = 0.739$ with Arena CW). Our findings show that measuring associative distance offers a highly effective way to assess LLMs’ creativity, capturing their ability to move beyond surface-level co-occurrence and tap into deeper, less common semantic connections that underlie genuine creativity.

Further analysis show that while high-performing LLMs match general human scores, professionals consistently outperform models and display more diverse associative patterns. These findings underscore both advances and limitations in LLM creative association, and highlight PACE as an effective tool for benchmarking and advancing model creativity.

% Compared to existing creativity evaluation methods, PACE not only avoids issues of training data contamination, but also provides a simple and scalable assessment framework that significantly reduces the need for costly manual evaluation. Further comparisons between LLMs and humans reveal that high-performing LLMs achieve scores comparable to the average human, while professional humans consistently outperform LLMs and exhibit greater diversity in associative patterns.

% These results highlight both the progress and current limitations of LLMs in creative association, and suggest that human creativity is characterized not only by higher performance but also by richer and more diverse associative strategies. PACE thus provides a valuable tool for benchmarking LLM creativity and guiding future model development.

\section*{Limitations}
\textbf{Limited focus on English.} Since we use English seed words and rely primarily on leaderboards with English-based rankings (such as Arena CW), the evaluation of PACE is conducted in English, focusing on its correlation with creativity performance. Consequently, our results are limited to the assessment of English creative ability.

\textbf{Limited sample model sizes.} To indirectly validate robustness, we rely on rankings from external leaderboards; however, this approach inherently constrains model selection due to the finite number of models represented across these platforms. Furthermore, to maintain comparability across disparate leaderboards, we restrict our analysis to models that consistently appear across all evaluated platforms, thereby further limiting the scope of our analytical sample. Following the guidelines established by \citet{bonett2000sample}, Spearman correlations within the moderate range of \(|\rho| \approx 0.5 - 0.7\) necessitate a minimum sample size of 20-30 observations to establish statistically reliable confidence intervals. 
% While our primary analyses of Arena All and Arena CW incorporate a sufficient number of models to meet these statistical requirements, several other leaderboard comparisons approach the lower boundary of this threshold, potentially introducing minor limitations to the robustness and generalizability of our findings.
While our primary analyses on Arena All and Arena CW include a sufficient number of models to ensure statistical reliability, some other leaderboard comparisons are closer to the minimum threshold, which may introduce minor limitations to the robustness and generalizability of our findings.

\section*{Acknowledgments}
The authors would like to thank Dr. Wang Yin and his lab members for their helpful discussions and suggestions. This research was partially supported by the Tencent Basic Platform Technology Rhino-Bird Focused Research Program.

% \bibliographystyle{acl_natbib}
\bibliography{main}

\clearpage
\appendix
\label{sec:appendix}
\section*{Appendices}
\section{Dataset Details}
\label{sec:selected_seed_words}
The final set of 110 seed words is selected through a two-step process. First, using the NLTK part-of-speech tagger, we identify nouns by filtering for words with the "NN" prefix, as nouns frequently serve as stimuli in association experiments. While our initial focus is on nouns, we include all identified words in our dataset since words from different syntactic categories can effectively trigger associations. Second, we rank these words based on their frequency in COCA2020 \cite{davies2008coca}, divide the corpus into five equal segments, and select the final words based on this stratification.

\begin{table}[h]
\small
\begin{adjustbox}{max width=\columnwidth}
\begin{tabular}{lp{0.7\columnwidth}}
\hline
\textbf{Chapter} & \textbf{Seed} \\
\hline
The physical world & rock, wood, dust, rainbow, headland \\
Kinship & son, female, widow, son-in-law, stepdaughter \\
Animals & eagle, worm, dove, firefly, midge \\
The body & sick, toe, blink, eyelid, earwax \\
Food and drink & meal, pepper, crush, ripe, unripe \\
Clothing and grooming & spin, soap, bracelet, braid, awl \\
The house & bed, pole, ladder, chimney, cookhouse \\
Agriculture and vegetation & grass, mushroom, bamboo, sickle, banyan \\
Basic actions and technology & strike, broken, cord, glue, adze \\
Motion & push, lift, swim, dive, outrigger \\
Possession & seek, hire, possess, lend, stingy \\
Spatial relations & center, ball, collect, round, fathom \\
Quantity & piece, count, pair, twelve, multitude \\
Time & month, summer, yesterday, cease, timepiece \\
Sense perception & dark, dry, rough, sour, brackish \\
Emotions and values & pain, correct, anxiety, sadness, deceit \\
Cognition & seem, explain, reflect, wise, imitate \\
Speech and language & speak, refuse, confess, howl, rebuke \\
Social and political relations & subject, neighbor, plot, ruler, chieftain \\
Warfare and hunting & peace, defeat, bow, fortress, fishhook \\
Law & murder, judgment, punishment, plaintiff, arson \\
Religion and belief & pray, temple, fairy, phantom, portent \\
\hline
\end{tabular}
\end{adjustbox}
\caption{Chapters and their associated seed words}
\end{table}

\section{Experimental Setup Details}
\subsection{Selected Models} 
Full list of selected models can be found in Table~\ref{tab:selected_models}. PACE evaluation contains a comprehensive selection of LLMs, featuring both leading open-source models (such as DeepSeek, Gemma, LLaMA, and Qwen series) and prominent closed-source commercial models (including various versions of Claude, Gemini, and GPT series). This balanced selection represents the current state-of-the-art across commercial and open-source domains, with a total of 34 models evaluated. Among these 34 models, 30 have corresponding Chatbot Arena Leaderboard scores and rankings (based on the early May 2025 scoring version, \citealp{chiang2024chatbot}), while the remaining four models (Command-R-Plus-08-2024, DeepSeek-R1-Distill-LLaMA-70b, DeepSeek-R1-Distill-Qwen-32b, and Hunyuan-Turbos-20250313) are included to ensure comprehensive coverage across different leaderboards, despite lacking Arena recordings.

\begin{table}[t]
\small
\begin{adjustbox}{max width=\columnwidth}
\begin{tabular}{lccc}
\hline
\textbf{Model} & \textbf{License} & \textbf{Arena CW} & \textbf{Association Distance} \\
\hline
gemini-2.5-pro-preview-03-25 & - & 1450 & 0.7757 \\
deepseek-chat-v3-0324 & \checkmark & 1376 & 0.7628 \\
gpt-4.1-2025-04-14 & - & 1364 & 0.7728 \\
deepseek-r1 & \checkmark & 1356 & 0.7588 \\
gemini-2.0-flash-001 & - & 1348 & 0.7576 \\
qwen3-235b-a22b & \checkmark & 1314 & 0.7553 \\
gemma-3-27b-it & \checkmark & 1358 & 0.7673 \\
qwen-max-2025-01-25 & - & 1334 & 0.7505 \\
deepseek-v3 & \checkmark & 1331 & 0.7480 \\
o3-mini-2025-01-31 & - & 1270 & 0.7388 \\
claude-3.7-sonnet & - & 1316 & 0.7817 \\
yi-lightning & - & 1282 & 0.7614 \\
claude-3.5-sonnet & - & 1289 & 0.7885 \\
gpt-4o-mini-2024-07-18 & - & 1270 & 0.7297 \\
gpt-4.1-nano & - & 1256 & 0.7340 \\
hunyuan-standard & - & 1244 & 0.7171 \\
llama-3.1-405b-instruct & \checkmark & 1264 & 0.7521 \\
llama-3.3-70b-instruct & \checkmark & 1255 & 0.7542 \\
qwen2.5-72b-instruct & \checkmark & 1228 & 0.7339 \\
mistral-large-2407 & \checkmark & 1246 & 0.7429 \\
mistral-large-2411 & \checkmark & 1246 & 0.7548 \\
llama-3.1-70b-instruct & \checkmark & 1239 & 0.7476 \\
gemma-2-27b-it & \checkmark & 1245 & 0.7488 \\
llama-3-70b-instruct & \checkmark & 1214 & 0.7532 \\
claude-3-sonnet & - & 1188 & 0.7345 \\
qwen2-72b-instruct & \checkmark & 1184 & 0.7371 \\
claude-3-haiku & - & 1163 & 0.7236 \\
mixtral-8x22b-instruct & \checkmark & 1147 & 0.7515 \\
gpt-3.5-turbo-0125 & - & 1099 & 0.7283 \\
gpt-3.5-turbo-1106 & - & 1044 & 0.7226 \\
command-r-plus-08-2024 & \checkmark & - & 0.7397 \\
deepseek-r1-distill-llama-70b & \checkmark & - & 0.7461 \\
deepseek-r1-distill-qwen-32b & \checkmark & - & 0.7437 \\
hunyuan-turbos-20250313 & - & - & 0.7260 \\
\hline
\end{tabular}
\end{adjustbox}
\caption{Selected Models with Arena CW Scores (Cutoff: Early May 2025) and Their Association Distances}
\label{tab:selected_models}
\end{table}

\subsection{Formula for Association Distance}  
\label{sec:formula_for_association_distances}
Our association distance measurement builds upon \citet{gray2019forward}. For each position $n$ in an association chain, we calculate the association distance as the average semantic distance from the current position to all preceding positions:

\begin{equation}
    A_n = \frac{\sum_{i=1}^{n-1} D_{n,i}}{n-1},
    \label{eq:instantaneous_assoc}
\end{equation}

where $D_{n,i}$ represents the semantic distance between positions $n$ and $i$, capturing the conceptual relatedness between thoughts at these positions.

The association distance of an entire sequence is then calculated by averaging the association distances across all positions:

\begin{equation}
    A_{\text{chain}} = \frac{\sum_{i=2}^{n} \left( \frac{\sum_{j=1}^{i-1} D_{i,j}}{i-1} \right)}{n-1},
    \label{eq:dynamic_assoc}
\end{equation}

where $n$ is the total number of positions in the association chain.

To enhance diversity of LLMs' responses, we generate three association chains for each seed. The association distance for each seed is computed by averaging the three chain scores:

\begin{equation}
    A_{\text{seed}} = \frac{1}{3}\sum_{c=1}^{3} A_{\text{chain},c},
    \label{eq:seed_assoc}
\end{equation}

Finally, the overall association distance metric for a model is derived by averaging across all seeds:

\begin{equation}
    A_{\text{model}} = \frac{1}{S}\sum_{s=1}^{S} A_{\text{seed},s},
    \label{eq:model_assoc}
\end{equation}

where $S$ represents the total number of seeds evaluated.

\subsection{Prompts} 
\label{sec:prompts}
We use a two-step approach to construct parallel association chains. First, we generate prompts based on the methodology proposed by \citet{gray2019forward}, incorporating more detailed instructions to articulate task requirements clearly. This modification addresses our observation that certain lower-tier language models tend to generate associations consistently based on the seed word rather than the immediately preceding word. Additionally, we require models to provide reasoning for each association between consecutive words, which serves two purposes: ensuring adherence to task specifications and enhancing label accuracy in association type classification. 

To compare different LLMs, we set the temperature parameter to 0 to observe their intrinsic associative patterns (with the exception of o3-mini, which has a fixed temperature setting of 1). 

\begin{tcolorbox}[
  title=First Stage Prompt,
  colback=blue!5,
  colframe=blue!50!black,
  width=\columnwidth,
  breakable,
  before skip=10pt,
  after skip=10pt,
  left=5pt,
  right=5pt]
Starting with the word "\{seed\}", generate three different words that directly associate with this initial word only (not with each other). Please put down only single words, and do not use proper nouns (such as names, brands, etc.). For each word, provide a brief explanation of its connection to "\{seed\}". Return in JSON format:
\begin{verbatim}
{
  "results": [
    {"word": "", "reason": ""},
    {"word": "", "reason": ""},
    {"word": "", "reason": ""}
  ]
}
\end{verbatim}
\end{tcolorbox}

\begin{tcolorbox}[
  title=Second Stage Prompt,
  colback=green!5,
  colframe=green!50!black,
  width=\columnwidth,
  breakable,
  before skip=10pt,
  after skip=10pt,
  left=5pt,
  right=5pt,
  fontupper=\small]

Starting with the word pair "\{seed\}" → "\{second\_word\}", generate a chain of 20 words where each new word should be associated with ONLY the word immediately before it. Generate the third word based on "\{second\_word\}", then generate the fourth word based on your third word, and so on. Please put down only single words, and do not use proper nouns (such as names, brands, etc.). For each word, provide a brief explanation of its connection to the previous word. Return in JSON format with exactly 20 entries:
\begin{verbatim}
{
  "results": [
    {"word": "{second_word}", 
    "reason": "{second_word_reason}"},
    {"word": "", "reason": ""},
    {"word": "", "reason": ""},
    ...
    {"word": "", "reason": ""}
  ]
}
\end{verbatim}
\end{tcolorbox}

\subsection{Settings for Comparison Between Human and LLMs} 
\label{sec:app_comparison_between_human_and_llms}
In Section~\ref{sec:comparison_between_human_and_llms}, we compare LLM and human performance using data from \citet{gray2019forward}. Specifically, we use two groups from the original study: "general" (representative American adults, Group 2 in the original paper) and "professional" (professional actors, Group 4 with actor label in the original paper). The professional group achieved the highest scores in both the original association task and the traditional psychological validation tests. We use all human data without additional cleaning.

For LLM analysis, we select two parallel groups based on their Arena All Rankings. The high-performing group comprises four LLMs ranked within the top 20: DeepSeek-Chat-v3.1, Gemini-2.5-Pro-03-25-preview, Qwen3-235b-a22b, and GPT-4.1. The mid-performing group includes Yi-Lightning, Gemma-2-27b-it, LlaMA-3.3-70b-Instruct, and Mistral-Large-2411, with an average ranking of 75 on the leaderboard, representing the standard performance of current models. 

For seed words, we use the same set from human studies: bear, table, candle, snow, paper, and toaster. To match human sample sizes, we vary LLM temperature between 0 and 1. We generate three independent association chains per seed word at each temperature setting, calculating metrics separately for each chain rather than averaging, thereby simulating multiple participants. Each model generates 6 chains per seed word (3 chains × 2 temperatures).

\section{Additional Experimental Results}
\label{sec:validation_results}
\subsection{Robustness Analyses}
\textbf{Correlation with different embedding models.} To validate the correlation, we employ three widely-used English word embeddings to compute association distances: GloVe (GloVe-6B-300d; \citealp{pennington2014glove}), MUSE (English; \citealp{conneau2017word}), and FastText (crawl-300d-2m; \citealp{mikolov2018advances}).

Results presented in Table~\ref{tab:spearman_results_with_different_embeddings} demonstrate a consistently significant correlation between PACE and Arena CW, with MUSE achieving the highest correlation coefficient ($\rho = 0.757$). To ensure consistency with the concreteness predictions, we choose FastText as the evaluation method. 

\begin{table}[ht]
\centering
\caption{Spearman Correlation Results Across Different Word Embedding Models}
\label{tab:spearman_results_with_different_embeddings}
\resizebox{\columnwidth}{!}{%
\begin{tabular}{lrrrrr}
\hline
\textbf{Leaderboard} & \textbf{Glove} & \textbf{Muse} & \textbf{FastText} & \textbf{Models} \\
\hline
Arena CW & 0.529**    & 0.757***    & 0.739***    & 30  \\
Arena All              & 0.488**    & 0.675***    & 0.660***    & 30  \\
MMLU-Pro               & 0.383      & 0.555**     & 0.505*      & 23  \\
Livebench              & 0.490*     & 0.651***    & 0.691***    & 19  \\
EQ-Bench               & 0.304      & 0.796***    & 0.637**     & 18  \\
\hline
\multicolumn{5}{l}{\small{* $p < 0.05$, ** $p < 0.01$, *** $p < 0.001$}} \\
\end{tabular}%
}
\end{table}

\textbf{Bootstrap results of correlation analysis.} To validate the robustness of the correlation coefficient, we use a bootstrap method to resample the results of seed words and compute the Spearman correlation. Table~\ref{tab:bootstrap_results} shows a stable and significant correlation with PACE rankings across all leaderboards (with a significance ratio of 1.000), with the exception of MMLU-Pro (which had a significance ratio of 0.962). Among these, Arena CW shows the strongest relationship, achieving the highest correlation with PACE, with Spearman correlation values ranging from 0.678 to 0.769.

\begin{table}[ht]
\centering
\caption{Bootstrap Results for Spearman Correlation Across Different Leaderboards}
\label{tab:bootstrap_results}
\resizebox{\columnwidth}{!}{%
\begin{tabular}{lrrrrr}
\hline
\textbf{Leaderboard} & \textbf{Mean Corr.} & \textbf{SE} & \textbf{95\% CI} & \textbf{Sig. Ratio} \\
\hline
Arena CW & 0.726*** & 0.023 & [0.678, 0.769] & 1.000 \\
Arena All & 0.650*** & 0.023 & [0.602, 0.695] & 1.000 \\
MMLU-Pro & 0.489*   & 0.045 & [0.405, 0.578] & 0.962 \\
LiveBench & 0.669*** & 0.031 & [0.607, 0.725] & 1.000 \\
EQ-Bench & 0.624**  & 0.043 & [0.537, 0.714] & 1.000 \\
\hline
\multicolumn{5}{l}{\small{* $p < 0.05$, ** $p < 0.01$, *** $p < 0.001$}} \\
\end{tabular}%
}
\end{table}

\textbf{Impact of reduced elements on correlation.} To enhance evaluation efficiency, we explore the impact of two parameters: the \textbf{number of seed words} (See Table~\ref{tab:spearman_correlations_num}) and the \textbf{chain length} (See Table~\ref{tab:spearman_correlations_length}). We use random sampling with 500 iterations to select various subsets of seed words. we also analyze the effect of different chain length by truncating the original chains and computing the correlation coefficients.

\begin{table}[ht]
\centering
\caption{Impact of Reducing Seed Nums}
\label{tab:spearman_correlations_num}
\resizebox{\columnwidth}{!}{%
\begin{tabular}{lrrrr}
\hline
\textbf{Leaderboard} & \textbf{Num-1} & \textbf{Num-2} & \textbf{Num-3} & \textbf{Num-4} \\
\hline
Arena CW     & 0.587 (0.048) & 0.609 (0.034) & 0.613 (0.025) & 0.617 (0.017) \\
Arena All    & 0.598 (0.050) & 0.621 (0.035) & 0.626 (0.025) & 0.630 (0.019) \\
MMLU-Pro     & 0.439 (0.084) & 0.453 (0.056) & 0.465 (0.045) & 0.471 (0.033) \\
LiveBench    & 0.589 (0.071) & 0.604 (0.051) & 0.613 (0.034) & 0.612 (0.027) \\
EQ-Bench     & 0.649 (0.080) & 0.673 (0.056) & 0.681 (0.040) & 0.686 (0.027) \\
\hline
\multicolumn{5}{l}{\small{* $p < 0.05$, ** $p < 0.01$, *** $p < 0.001$}} \\
\end{tabular}%
}
\end{table}

\begin{table}[ht]
\centering
\caption{Impact of Reducing Chain Length}
\label{tab:spearman_correlations_length}
\resizebox{\columnwidth}{!}{%
\begin{tabular}{lrrrr}
\hline
\textbf{Leaderboard} & \textbf{Length-5} & \textbf{Length-10} & \textbf{Length-15} & \textbf{Length-20} \\
\hline
Arena CW & 0.582*** & 0.698*** & 0.717*** & 0.739*** \\
Arena All & 0.502** & 0.618*** & 0.637*** & 0.660*** \\
MMLU-Pro & 0.249  & 0.479*  & 0.461*  & 0.505*  \\
LiveBench & 0.558*  & 0.632** & 0.633** & 0.691** \\
EQ-Bench & 0.370  & 0.554*  & 0.562*  & 0.637** \\
\hline
\multicolumn{5}{l}{\small{* $p < 0.05$, ** $p < 0.01$, *** $p < 0.001$}} \\
\end{tabular}%
}
\end{table}

The results demonstrate that larger sample sizes yield higher correlation coefficients, indicating enhanced performance stability. 

\subsection{Extended Results on Models and Human Performance}
\begin{figure*}[htbp]
\centering
\includegraphics[width=\textwidth]{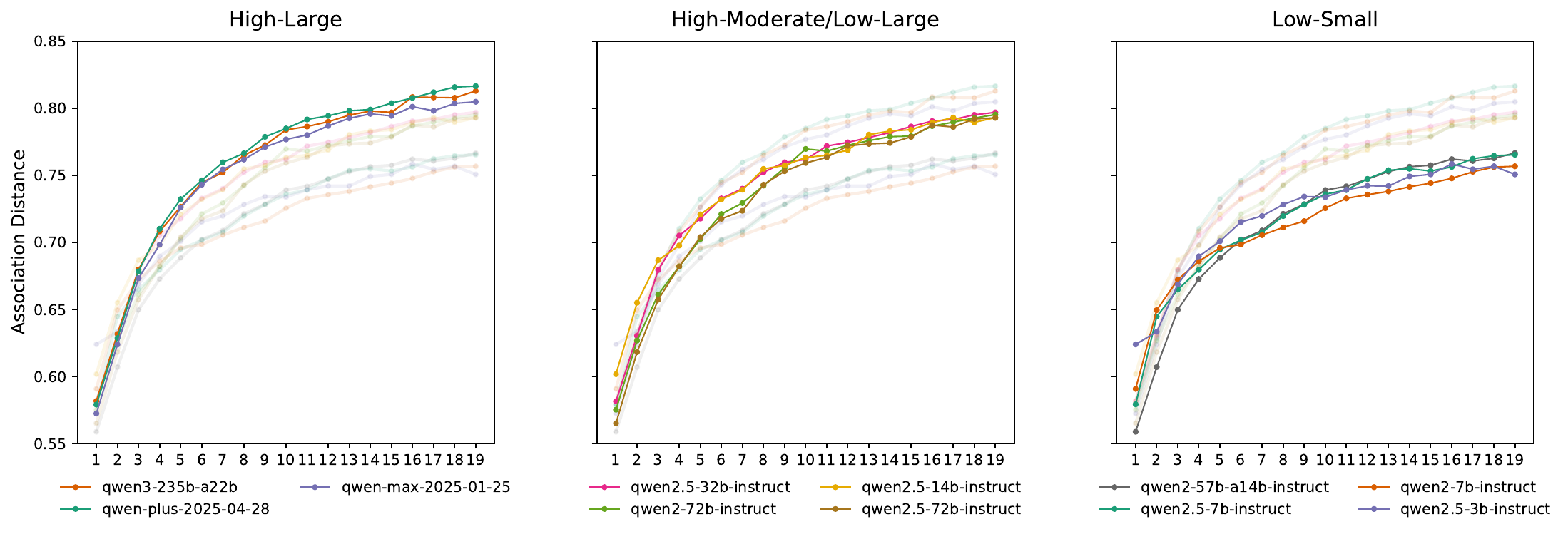}
\caption{\textbf{Association distance comparison across versions and sizes of Qwen models.} This figure represents the association distance calculated at each position within the associative chains across different models and versions. Results reveal three performance clusters at different chain positions: (1) \textbf{high-large models} (new architectures, larger parameters), (2) \textbf{high-moderate and low-large models} (mixed newer models with moderate parameters and older models with larger parameters), and (3) \textbf{low-small models} (smaller architectures, fewer parameters). These findings highlight the combined effect of model version and parameter size and validate PACE as an effective evaluation framework.}
\label{fig:association_distance_comparison}
\end{figure*}

\begin{figure*}[htbp]
\centering
\includegraphics[width=\textwidth]{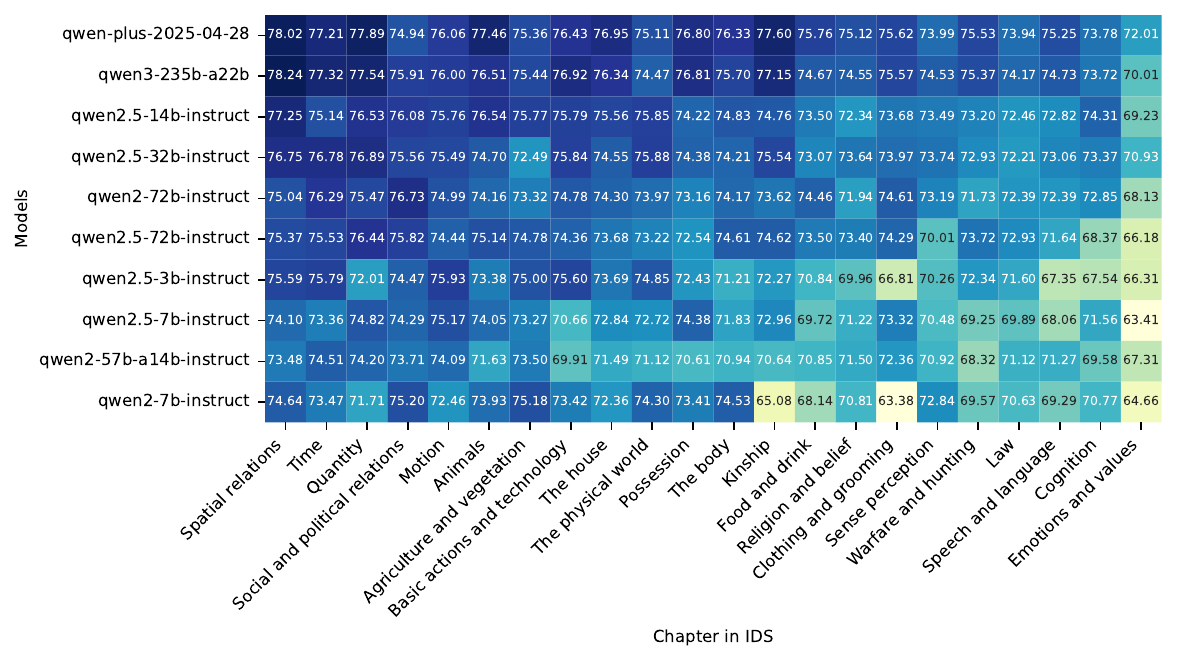}
\caption{Association Distance Sorted by Chapters in IDS.  
The heatmap presents the association distance scores of different Qwen model versions across 22 semantic chapters in the IDS dataset. Each cell represents the performance score of a model (rows) on a particular semantic category (columns), with darker shades indicating higher scores. Objective categories, such as \textbf{Spatial relations}, \textbf{Time}, and \textbf{Quantity}, show consistently high performance across models, whereas subjective and abstract categories, such as \textbf{Kinship} and \textbf{Emotions and values}, display larger performance gaps, highlighting improvements in newer models.}
\label{fig:category_heatmap_qwen}
\end{figure*}

\textbf{Model comparisons.}
\label{sec:pace_qwen_models} 
We compare different Qwen models to investigate how model versions and sizes influence association distances. As shown in Figure~\ref{fig:association_distance_comparison}, association scores consistently follow a hierarchical pattern by version: Qwen-3 outperforms Qwen-2.5, which in turn outperforms Qwen-2. While smaller models from various versions (e.g., Qwen-2-7b, Qwen-2.5-3b) generally fall into lower-performing groups, larger models from older versions can still achieve performance comparable to that of more recent releases (e.g., Qwen-2-72b vs. Qwen-2.5-14b).  
% These findings indicate that architectural improvements and increased model size serve as distinct yet complementary approaches to enhancing model performance.
% the association scores differ clearly across model series, following a consistent hierarchy by version: Qwen-3, Qwen-2.5, and Qwen-2. In addition, although smaller models from various versions (e.g., Qwen-2-7b, Qwen-2.5-3b) usually fall into lower-performing groups, larger models from older versions can still match the performance of newer releases (e.g., Qwen-2-72b). This result highlights that architectural advances and increased parameters represent distinct but complementary strategies for improving model performance.

\textbf{Performance across semantic categories.} 
We further examine how semantic categories influence performance across different model versions. While newer models generally outperform older ones, the gap varies considerably by category. Strong performance is observed in objective categories such as \textbf{spatial relations, time, and quantity}, where even the earliest, smallest model (qwen-2-7b) achieves scores above 0.71. In contrast, subjective and abstract categories (e.g., \textbf{emotions and values}, 0.63–0.72; \textbf{kinship}, 0.65–0.78) exhibit substantially larger gaps, with newer models achieving up to 0.10 higher scores compared to prior versions.

\textbf{Human–LLM lexical diversity.} 
We combine responses from humans and LLMs and standardize sample sizes for each seed word to eliminate potential biases from differing data amounts. Type-Token Ratio (TTR) analysis reveal clear differences in lexical diversity: even when prompted and configured with temperatures of 0 and 1 to enhance diversity (see Sections~\ref{sec:parallel_word_association} and~\ref{sec:app_comparison_between_human_and_llms}), LLMs consistently exhibit lower TTR values than humans across all seed words. This suggests that LLMs produce more homogeneous responses, highlighting their limitations as substitutes for human creative output \cite{walsh2024does, wenger2025we}.

\textbf{Examples of association chains across the score spectrum.} 
Table~\ref{tab:semantic_distance_examples} shows responses from humans and LLMs given the same seed word \textbf{candle}. Human responses tend to exhibit more jumping associations (e.g., lucky → irish → friend → wedding), while model-generated responses are generally more uniform and sequential (e.g., liquid → water → rain → storm).

\begin{figure}[t]
    \centering
    \includegraphics[width=8cm]{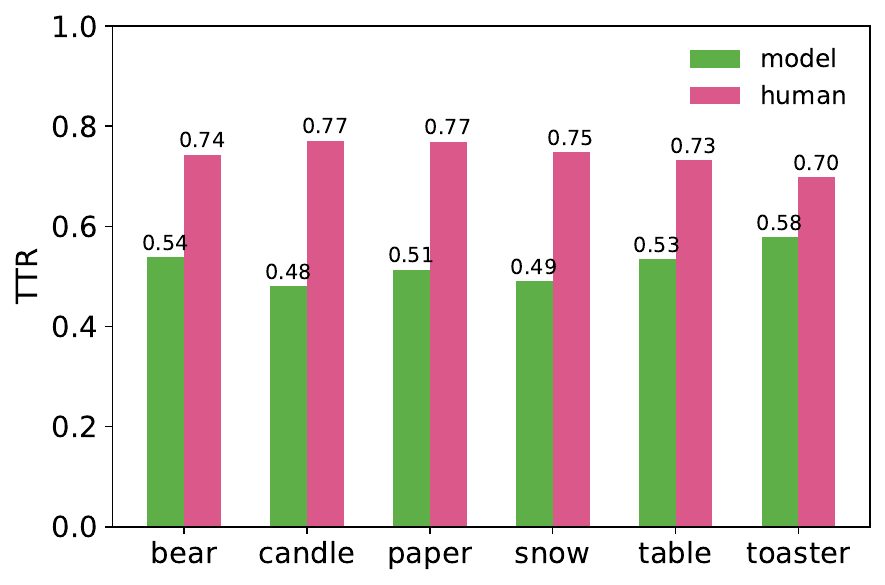}
    \caption{Type–Token Ratio (TTR) of responses generated by humans and models}
    \label{fig:ttr}
\end{figure}

\begin{table*}[t]
\centering
\caption{Examples of Association Chains with Different Semantic Distances}
\label{tab:semantic_distance_examples}
\begin{tabularx}{\textwidth}{p{3cm} X p{2cm}}
\toprule
\textbf{Source} & \textbf{Association Chain} & \textbf{Score} \\
\midrule
professional & candle, flame, basketball, dad, lucky, irish, friend, wedding, ring, run, air, cold, ski, sister, proud, lion, documentary, satire, podcast, subway & 0.8300 \\
\addlinespace
general & candle, fire, water, swim, kids, family, love, marriage, commitment, honor, life, decisions, problems, solutions, work, play, fun, joy, pain, death & 0.8060 \\
\addlinespace
gemini-2.5-pro & candle, blackout, darkness, night, sleep, dream, fantasy, story, book, paper, tree, forest, wildlife, nature, growth, plant, seed, potential, energy, force & 0.7947 \\
\addlinespace
llama-3.3-70b & candle, flame, heat, burn, fire, danger, warning, sign, symbol, language, communication, message, letter, paper, tree, forest, wildlife, habitat, ecosystem, balance & 0.7651 \\
\addlinespace
mixtral-large-2411 & candle, wick, flame, heat, melt, liquid, water, rain, storm, lightning, thunder, noise, silence, calm, serene, peaceful, tranquil, relax, sleep, dream & 0.7112 \\
\bottomrule
\end{tabularx}
\end{table*}

\section{Additional Discussion}
\subsection{Association Type Classification}
\label{sec:labelling_association_type}
Given that LLMs have demonstrated the ability to identify various types of associations \cite{de2024can}, we use DeepSeek-V3.1 to classify the semantic relationships between consecutive word pairs in each association chain. The classification followed the association type framework described by \citet{nissen2006word}, which distinguishes four categories: \textbf{paradigmatic} (same word class with semantic relations like synonymy/antonymy, e.g., \textit{love-heart}), \textbf{syntagmatic} (sequential/syntactic connections across word classes, e.g., \textit{local–politician}), \textbf{phonological} (sound-based similarity without semantic connection, e.g., \textit{quote-vote}), and \textbf{other} (personal associations, morphological variations, or unclassifiable connections, e.g., \textit{desperate-rhino}). Table~\ref{tab:word_association} presents examples of classified association types.

\subsection{Concreteness Prediction via Embedding Models} 
\label{sec:prediction_of_concreteness_using_embedding_models}
Word embeddings have been shown to effectively predict semantic concreteness and other psychological dimensions of words \cite{charbonnier2019predicting, flor2024three, hussain2024probing}. We use the concreteness dataset (40,000+ words) developed by \citet{brysbaert2014concreteness}, one of the largest human-labeled concreteness databases, to train concreteness prediction models using three different word embedding approaches: FastText (English), GloVe (6B-300d), and MUSE (English). Model performance is evaluated using Pearson's correlation coefficient, root mean square error (RMSE), and Kendall's rank correlation. Table~\ref{tab:embeddings_comparison} indicates that FastText achieves the highest Pearson correlation and Kendall coefficient, as well as the lowest RMSE. Therefore, we use FastText to assign concreteness ratings to association responses. Table~\ref{tab:word_association} shows examples of concreteness ratings assigned using this approach.

\begin{table}[H]
\centering
\caption{Comparison of word embedding models for concreteness prediction}
\label{tab:embeddings_comparison}
\resizebox{\columnwidth}{!}{%
\begin{tabular}{lccc}
\hline
\textbf{Model} & \textbf{Pearson } $r$ & \textbf{Kendall } $\tau$ & \textbf{RMSE} \\
\hline
\multicolumn{4}{l}{\textit{Training Set}} \\
FastText & \textbf{0.931 ± 0.000} & \textbf{0.760 ± 0.001} & \textbf{0.371 ± 0.001} \\
GloVe & 0.902 ± 0.001 & 0.728 ± 0.001 & 0.442 ± 0.001 \\
MUSE & 0.848 ± 0.001 & 0.658 ± 0.001 & 0.541 ± 0.001 \\
\hline
\multicolumn{4}{l}{\textit{Test Set}} \\
FastText & \textbf{0.910 ± 0.002} & \textbf{0.722 ± 0.003} & \textbf{0.421 ± 0.004} \\
GloVe & 0.837 ± 0.004 & 0.638 ± 0.004 & 0.556 ± 0.006 \\
MUSE & 0.845 ± 0.004 & 0.654 ± 0.004 & 0.545 ± 0.005 \\
\hline
\end{tabular}%
}
\small \raggedright \textit{Note:} Values shown as mean ± standard deviation. \textbf{Bold} indicates best performance. Valid words: FastText (35,424), GloVe (31,617), MUSE (27,101).
\end{table}

\begin{table}[H]
\centering
\caption{Fixed Effects Model Results Across Groups}
\label{tab:fixed_effects}
\resizebox{\columnwidth}{!}{%
\begin{tabular}{lrrr}
\hline
\textbf{Group} & \textbf{Intercept ($\beta_0$)} & \textbf{Slope ($\beta_1$)} & \textbf{R\textsuperscript{2}} \\
\hline
mid-llm & 4.295 & -0.023*** & 0.249 \\
high-llm & 4.197 & -0.020*** & 0.279 \\
general & 4.022 & -0.025*** & 0.350 \\
professional & 3.987 & -0.017** & 0.275 \\
\hline
\end{tabular}%
}
\small \raggedright
\vspace{3pt}
\textit{Note:} *** $p<0.001$, ** $p<0.01$. 95\% confidence intervals in brackets. Responses shorter than 20 words were excluded from analysis.
\end{table}

\begin{table*}[t]
\centering
\caption{Examples of Concreteness and Association Types}
\label{tab:word_association}

% (a) High-LLM (GPT-4.1)
\small
\textbf{(a) High-LLM (GPT-4.1)}

\vspace{0.2cm}
\setlength{\tabcolsep}{4pt} 
\begin{tabular}{l|>{\centering\arraybackslash}p{1.2cm}|*{9}{>{\centering\arraybackslash}p{1.2cm}}}

\toprule
\textbf{Pos} & \textbf{Seed} & \textbf{2} & \textbf{3} & \textbf{4} & \textbf{5} & \textbf{6} & \textbf{7} & \textbf{8} & \textbf{9} & \textbf{10} \\
\midrule
Word & toaster & appliance & kitchen & cooking & heat & fire & wood & tree & forest & wildlife \\
Type & -- & para & syn & syn & syn & syn & syn & para & para & syn \\
Conc. & -- & 4.31 & 4.86 & 3.96 & 3.75 & 4.54 & 5.00 & 4.92 & 4.76 & 4.17 \\
\bottomrule
\end{tabular}

\vspace{0.2cm}

\setlength{\tabcolsep}{4pt} 
\begin{tabular}{l|*{10}{>{\centering\arraybackslash}p{1.2cm}}}
\toprule
\textbf{Pos} & \textbf{11} & \textbf{12} & \textbf{13} & \textbf{14} & \textbf{15} & \textbf{16} & \textbf{17} & \textbf{18} & \textbf{19} & \textbf{20} \\
\midrule
Word & animal & mammal & fur & coat & winter & snow & flake & crystal & glass & window \\
Type & para & para & syn & syn & syn & syn & para & para & syn & syn \\
Conc. & 4.63 & 4.87 & 4.63 & 4.97 & 3.87 & 5.00 & 4.18 & 4.39 & 5.00 & 4.68 \\
\bottomrule
\end{tabular}

\vspace{0.4cm}

% (b) Professional group
\small
\textbf{(b) Professional group}
\vspace{0.2cm}

\setlength{\tabcolsep}{4pt} 
\begin{tabular}{l|>{\centering\arraybackslash}p{1.2cm}|*{9}{>{\centering\arraybackslash}p{1.2cm}}}
\toprule
\textbf{Pos} & \textbf{Seed} & \textbf{2} & \textbf{3} & \textbf{4} & \textbf{5} & \textbf{6} & \textbf{7} & \textbf{8} & \textbf{9} & \textbf{10} \\
\midrule
Word & toaster & bread & money & struggle & tussle & tout & flout & flounce & bounce & bunny \\
Type & -- & syn & other & other & para & pho & pho & pho & pho & pho \\
Conc. & -- & 4.92 & 3.46 & 2.36 & 3.12 & 2.62 & 2.41 & 3.39 & 3.79 & 4.77 \\
\bottomrule
\end{tabular}

\vspace{0.2cm}

\setlength{\tabcolsep}{4pt} 
\begin{tabular}{l|*{10}{>{\centering\arraybackslash}p{1.2cm}}}
\toprule
\textbf{Pos} & \textbf{11} & \textbf{12} & \textbf{13} & \textbf{14} & \textbf{15} & \textbf{16} & \textbf{17} & \textbf{18} & \textbf{19} & \textbf{20} \\
\midrule
Word & funny & laughter & song & dance & jig & pig & fortune & fame & frame & lame \\
Type & pho & para & syn & syn & para & pho & other & other & pho & pho \\
Conc. & 2.57 & 3.72 & 3.96 & 4.26 & 4.27 & 5.00 & 3.04 & 2.39 & 4.19 & 2.43 \\
\bottomrule
\end{tabular}

\vspace{0.2cm}
\parbox{\textwidth}{\raggedright
\textit{Note:} Pos = Position; Conc. = Concreteness; syn = syntagmatic; para = paradigmatic; pho = phonological; other = other types. 
Association types are labeled by DeepSeek-V3.1 and concreteness scores are predicted by FastText based on data from \citet{brysbaert2014concreteness}. 
The original concreteness scores range from 0 to 5; predictions exceeding this range are clipped to the nearest boundary value.
}
\end{table*}

\end{document}